\documentclass[lettersize,journal]{IEEEtran}
\usepackage{amsmath,amsfonts}
\usepackage{algorithmic}
\usepackage{algorithm}
\usepackage{array}
\usepackage[caption=false,font=normalsize,labelfont=sf,textfont=sf]{subfig}
\usepackage{textcomp}
\usepackage{stfloats}
\usepackage{url}
\usepackage{verbatim}
\usepackage{graphicx}
\usepackage{cite}

\begin{document}

\title{A Cross-Domain Few-Shot Learning Method\\  Based on Domain Knowledge Mapping}

\author{Jiajun Chen, Hongpeng Yin, Yifu Yang
\thanks{Jiajun Chen, Hongpeng Yin are with the School of Automation, Chongqing University, Chongqing, 400044, China. Yifu Yang is with the Bradley Department of Electrical
\& Computer Engineering, Virginia Polytechnic Institute and State University, Blacksburg, VA, USA 24061. (email: chenjiajun@stu.cqu.edu.cn; yinhongpeng@cqu.edu.cn; cyang826@vt.edu).}}



\maketitle

\begin{abstract}
In task-based few-shot learning paradigms, it is commonly assumed that different tasks are independently and identically distributed (i.i.d.). However, in real-world scenarios, the distribution encountered in few-shot learning can significantly differ from the distribution of existing data. Thus, how to effectively leverage existing data knowledge to enable models to quickly adapt to class variations under non-i.i.d. assumptions has emerged as a key research challenge. To address this challenge, this paper proposes a new cross-domain few-shot learning approach based on domain knowledge mapping, applied consistently throughout the pre-training, training, and testing phases. In the pre-training phase, our method integrates self-supervised and supervised losses by maximizing mutual information, thereby mitigating mode collapse. During the training phase, the domain knowledge mapping layer collaborates with a domain classifier to learn both domain mapping capabilities and the ability to assess domain adaptation difficulty. Finally, this approach is applied during the testing phase, rapidly adapting to domain variations through meta-training tasks on support sets, consequently enhancing the model’s capability to transfer domain knowledge effectively. Experimental validation conducted across six datasets from diverse domains demonstrates the effectiveness of the proposed method. 
\end{abstract}


\section{Introduction}
\IEEEPARstart{A}{t} present, benefiting from rapid advancements in computational power and model complexity, models trained on large-scale datasets using cross-entropy loss have achieved remarkable classification performance. However, it remains a significant challenge to adapt these pre-trained models to a completely different domain, with no overlap in distribution or categories, using only a minimal number of samples. This requires designing parameter learning strategies to endow models with rapid learning capabilities. Hence, this we investigates how models possessing extensive prior knowledge can utilize very few samples to quickly acquire learning ability in a completely new domain. Under this setting, not only is there a substantial distributional shift between the source and target domains, but also the category spaces satisfy the condition \( Y_S \cap Y_T = \emptyset \), meaning traditional transfer mechanisms based on category-related assumptions are no longer applicable.

We are built upon the meta-learning paradigm \cite{a120}. Distinct from category-based learning driven by cross-entropy, meta-learning focuses on tasks as learning objectives, extracting transferable knowledge from multiple tasks to rapidly adapt to new tasks. Its central idea is task modeling, allowing models to efficiently learn new tasks from very few samples. In other words, meta-learning excels at quickly generalizing models to novel classes under independent and identically distributed (i.i.d.) conditions. Although existing meta-learning methods such as MAML \cite{a120} and ProtoNet \cite{a114} have successfully provided models with rapid adaptation abilities through task-driven training paradigms, they assume tasks are sampled from the same meta-distribution. However, when faced with cross-domain scenarios, the local parameter updates obtained through inner-loop fine-tuning may converge to suboptimal solutions due to heterogeneity in underlying feature spaces. More critically, when target domain categories are completely unrelated to those in the source domain, traditional metric mechanisms based on prototypical or relational networks become ineffective, as they lack transferable semantic similarity criteria.

A considerable number of researchers have recognized the challenge posed by distribution shifts that cause existing meta-learning methods to become ineffective, prompting various related studies. Most current approaches attempt to build models within the meta-learning framework to rapidly adapt to domain variations. Tseng et al. \cite{a138} introduced a feature transformation layer, which can achieve domain adaptation for target tasks with only a few hyperparameters. However, when trained on a single source domain, these hyperparameters must be manually tuned, resulting in limited adaptability to domain shifts. Hu et al. \cite{a79} proposed a simulated distribution data augmentation method, employing adversarial training to emulate domain distributions at the data feature level. However, this approach is restricted to the meta-training phase and can only simulate distributional shifts within a limited range, making it difficult to ensure the validity of the simulated distributions. Similarly, Li et al. \cite{a139} presented a knowledge transformation method designed to mitigate domain shifts from training to testing. They constructed a feature transformation module based on sparse representation, transferring knowledge from the support set to the query set, and trained the model using self-supervised learning for consistent representation. However, this module’s scope is confined to the support and query sets within the same domain, limiting its capability to transform domain-specific knowledge.

The above methods focus primarily on leveraging limited known source-domain samples and employ various transformation mechanisms to map knowledge into the target domain, thereby enhancing classification performance. Additionally, other researchers have attempted to impose alternative constraints on models to promote generalization. Oh et al. \cite{a140} investigated the influence of different pre-training constraints on cross-domain few-shot learning. They explored the feature spaces produced by supervised and self-supervised pre-training under dimensions of domain similarity and few-shot transfer difficulty, examining their effects on further learning capacity under cross-domain few-shot scenarios. Their findings suggest that when domain similarity is low, self-supervised pre-training yields better transferability due to limited source domain information. Conversely, when domain similarity is high but sample difficulty is greater, supervised learning outperforms self-supervised learning, as it better handles challenging samples. It should be noted, however, that their method did not employ the meta-learning paradigm but instead used supervised training with \(20\%\) labeled samples in the target domain, measuring few-shot difficulty using accuracy metrics.

Yazdanpanah et al. \cite{a141} explored the impact of batch normalization on cross-domain few-shot learning, finding that fine-tuning affine parameters during the adaptation phase further enhances model performance. Zhou et al. \cite{a121} observed that prototypical networks rely on biased features to distinguish classes in meta-learning, causing them to fail under domain shifts. To mitigate this effect induced by data-domain biases, they proposed a local-global distillation model aimed at removing such biases.

At present, exploration in cross-domain few-shot learning is diverse, but still exhibits certain limitations. Researchers typically handle cross-domain issues by employing transformation modules or mapping mechanisms that primarily focus on the training phase. These approaches attempt to substitute domain shifts with task differences; however, differences among tasks often do not capture the properties associated with non-independent and identically distributed (non-i.i.d.) data. Consequently, models trained in this manner still face adaptation difficulties when encountering different data domains during the testing phase. Additionally, existing methods emphasize only the training stage, neglecting the critical impact that the pre-training phase has on the convergence direction of model parameters. These limitations will be specifically demonstrated in detail in the following subsection. To address the above issues, we propose a domain-knowledge mapping method. During the pre-training phase, in order to prevent category mode collapse, the proposed method integrates cross-entropy and self-supervised learning based on mutual information theory, encouraging the model to learn more generalized feature representations. This approach enables the classifier to rapidly adapt to significant domain-induced feature variations during meta-training tasks. In the training phase, the method assesses the transfer difficulty between visible and unseen domains, dynamically adjusting domain-knowledge mapping weights to suit tasks with varying degrees of transfer difficulty. Notably, the domain-knowledge mapping is applied not only during the training stage but also at the testing phase, where it quickly adapts to domain difficulty variations by updating mapping weights.

\section{Related Works}
The cross-domain few-shot learning scenario synthesizes two core challenges commonly observed in real-world applications: label space inconsistency stemming from long-tail distributions and domain distribution mismatch inherent to few-shot domain adaptation. To better reflect practical settings, this newly emerged complex problem—formed by the combination of label space inconsistency and domain shift—has become a focal point of current research. In recent years, studies in the field of cross-domain few-shot learning have predominantly concentrated on two directions: sample augmentation and feature transformation. These approaches aim to enhance knowledge transfer efficiency and improve cross-domain feature representations, thereby boosting the generalization ability of models. The following section provides a review of the current state of research from these two perspectives.

\textbf{Sample Augmentation}. Pseudo-label generation methods leverage unlabeled data by using model inference to generate supervisory signals, thereby expanding the effective information available for training. In early work, Yao et al. \cite{a69} integrated data augmentation with pseudo-labeling by applying rotation transformations to unlabeled samples and treating the rotation angle as a pseudo-label to construct a supervised task. This approach effectively transformed unlabeled data into a supervised learning setting. Subsequent studies introduced target-domain data during the pre-training phase. Phoo et al. \cite{a70} proposed incorporating target-domain samples into the pre-training stage, where a source-domain pre-trained model is used to generate high-confidence pseudo-labels for target-domain samples. This semi-supervised strategy retains the knowledge from the source domain while embedding prior information about the target domain’s feature distribution, enabling the model to gain an initial understanding of target-domain features. Further advancements were made by Samarasinghe et al. \cite{a71} in the domain of video action recognition, where temporal feature modeling and cross-domain pseudo-label optimization were employed to enhance the quality of pseudo-labels. Their approach significantly improved the robustness of few-shot video action classification. At present, the reliability of pseudo-labels directly influences model performance. The core challenge in pseudo-label generation lies in mitigating noisy labels and calibrating confidence scores. Current research typically addresses this by applying confidence threshold filtering and ensemble-based multi-model prediction strategies to suppress the impact of noisy labels.

Feature fusion techniques aim to enhance representation capabilities by integrating features extracted from different depths of neural networks, thereby mitigating overfitting caused by relying on single-level representations. Feature fusion is generally categorized into multi-level feature fusion and multi-model feature fusion. Multi-level feature fusion focuses on optimizing the representational characteristics across various layers of deep neural networks. Adler et al. \cite{a72} enhanced the model's sensitivity to high-level semantic features by computing classification loss from deeper layers and aggregating gradients accordingly. Zou et al. \cite{a73} introduced a mid-level residual prediction mechanism that preserves spatial details while strengthening the discriminative power of semantic representations. In contrast, multi-model feature fusion emphasizes integrating heterogeneous models to leverage the complementary representational strengths of different architectures. Representative approaches include dual-path structures that fuse semantic graph models with classifier outputs \cite{a74}, robust representation construction via multi-model feature averaging \cite{a75}, and feature decoupling and recombination through nonlinear subspace mapping \cite{a76}. Studies have shown that multi-level fusion is more effective for optimizing single models, while multi-model fusion is better suited for handling cross-modal or multi-source data scenarios.

Feature re-weighting techniques dynamically adjust the importance of features to enable adaptive optimization of models across different tasks or domain environments. Du et al. \cite{a77} proposed a meta-learning-based weight allocation mechanism, assigning learnable weight parameters independently to multi-level features and employing a meta-optimization process to achieve task-aware feature selection. Sa et al. \cite{a78} adopted an attention-guided re-weighting strategy, utilizing a multi-head attention module to capture contextual dependencies across hierarchical features and establish a fine-grained feature importance evaluation framework. Hu et al. \cite{a79} introduced adversarial training-driven weight optimization, where domain discrepancies were simulated through perturbations in the feature space, and an adversarial loss function was employed to encourage the model to focus on domain-invariant features. More recent work \cite{a80} has begun exploring the temporal dynamics of weight generation mechanisms, designing memory-augmented forget-update modules that allow the re-weighting process to evolve adaptively with different training stages. Overall, re-weighting strategies have transitioned from static rule-based weighting to dynamically learned adaptive weighting, significantly enhancing model flexibility in cross-domain scenarios.

\textbf{Feature Transformation and Alignment}. Feature transformation techniques achieve cross-domain adaptation by reshaping feature distributions through mathematical space-mapping methods, thereby enabling models to learn transferable feature representations across domains. Key approaches include geometric algebra-based spatial transformations that project features into high-dimensional geometric spaces to preserve cross-domain relational invariance \cite{a81}; adaptive distribution transformation strategies that dynamically adjust the shape of feature distributions via learnable transformation matrices \cite{a82}; and meta-learning-driven feature abstraction methods that generate meta-point representations with domain generalization capabilities for spatial conversion \cite{a83}. Wang et al. \cite{a84} integrated a memory-augmented transformation mechanism, leveraging external memory banks to store typical cross-domain feature patterns. These stored patterns are retrieved and recombined through a memory retrieval mechanism to reconstruct features and refine the feature space. While these methods significantly enhance the coverage of target domain distributions without compromising the discriminative power of source domain representations, challenges remain in reducing their computational complexity.

Feature alignment techniques focus on cross-domain distribution matching by minimizing domain discrepancies through explicit or implicit constraint mechanisms. Explicit alignment methods often adopt subspace projection strategies, mapping source and target domain features into a shared latent space to directly align their distributions. Guan et al. \cite{a85} employed Maximum Mean Discrepancy (MMD) to measure distributional distances, enforcing alignment between domains. Chen et al. \cite{a86} proposed a bidirectional alignment network that simultaneously optimizes the adaptation processes in both directions. In contrast, implicit alignment methods achieve distribution alignment indirectly by designing specific optimization objectives. Xu et al. \cite{a87} leveraged the principle of mutual information maximization to enhance the clustering properties of target domain features, while Guo et al. \cite{a88} corrected domain shifts by adjusting task-specific parameters. With the growing use of Transformer-based alignment architectures, which capture long-range feature dependencies through self-attention mechanisms, Paeedeh et al. \cite{a89} applied Transformer architectures for video-level cross-domain alignment. Zhao et al. \cite{a90} combined prototype alignment with distribution normalization to jointly optimize local sample matching and global distribution calibration.

Due to the dual inconsistency of label space and domain space inherent in cross-domain few-shot learning, it is difficult for any single method to effectively address this complex challenge. Recent studies have increasingly emphasized the integration of multiple approaches—such as combining pseudo-label generation with feature alignment, or jointly applying feature transformation and re-weighting strategies. Additionally, transfer strategies based on pre-trained models have introduced a new foundation for cross-domain few-shot learning. However, efficiently fine-tuning large-scale models under limited sample conditions remains an open research question. Future work may focus on constructing theoretical frameworks for domain generalization, enhancing adversarial robustness in few-shot scenarios, and enabling collaborative transfer of multimodal features.

\section{Problem Description}
According to the general formulation of meta-learning \cite{a121}, the primary goal of meta-learning is to extract transferable knowledge from a set of related tasks, enabling the model to rapidly adapt to new tasks using only a few samples. The meta-learning framework is built upon a task-driven learning paradigm, emphasizing generalization at the level of task distributions, rather than relying on the traditional assumption of independently and identically distributed (i.i.d.) samples in classical machine learning. In meta-learning, both training and testing revolve around tasks as fundamental units. Assume there exists a task distribution \( P(\tau) \), from which each task \( \tau_i \) is drawn. In a few-shot learning scenario, the dataset is partitioned into three subsets: a meta-training set, a meta-validation set, and a meta-test set, each containing multiple tasks.

Assume the sample space consists of an input space X and a label space Y, where label intersections across different tasks \( Y^{\tau} \) are empty sets. Each task \( \tau_i \) is further divided into a support set \( S^{\tau} = \{ (x_i^{\tau}, y_i^{\tau}) \}_{i=1}^{K \times N} \) containing N classes with K samples per class, and a query set \( Q^{\tau} = \{ (x_j^{\tau}, y_j^{\tau}) \}_{j=1}^{M} \) used to evaluate the model performance after task adaptation. Following the setting of \((N)\text{-}way\ (K)\text{-}shot\), the support set consists of N classes, each class containing K samples, which are used for task-specific parameter adaptation. The query set includes additional samples from the same classes and is used to evaluate the performance of the adapted model. This division compels the model to capture the essential characteristics of tasks using very few samples, thereby avoiding overfitting to specific samples.

During meta-training, for each task \( \tau_i \), the model starts from meta-parameters \( \theta \), computes the loss using the support set \( S^{\tau} \), and updates parameters via gradient descent as follows:

\begin{equation}
\label{deqn_ex1a}
\theta_i' = \theta - \alpha \nabla_{\theta} \mathcal{L}_{S^{\tau}}(\theta)
\end{equation}

where \( \alpha \) denotes the learning rate, and \( \mathcal{L}_{S^{\tau}}(\theta) \) represents the loss function.

Based on the losses computed across all tasks on the query sets \( Q^{\tau} \), the meta-parameters \( \theta \) are updated to minimize the generalization error after adaptation:

\begin{equation}
\label{deqn_ex1a}
\theta \leftarrow \theta - \beta \nabla_{\theta} \sum_{T_i} \mathcal{L}_{Q^{\tau}}(\theta_i')
\end{equation}

where \( \beta \) is the learning rate for the meta-query set. This optimization objective forces the meta-parameters \( \theta \) to initialize at a meta-initialization point sensitive to the task distribution, enabling rapid convergence to an optimal solution for new tasks within a few gradient steps. The gradient computation involves second-order derivatives \( \nabla_{\theta_i'} \), requiring explicit calculation of the Hessian matrix.

The effectiveness of meta-learning relies on the assumption of task distribution similarity, meaning that the meta-training tasks and meta-testing tasks share an underlying data generation mechanism. However, under the cross-domain few-shot learning scenario, the meta-training set and meta-testing set originate from distinct distributions. The training data domain is referred to as the source domain, while the unseen test domain is termed the target domain, that is, \( P_{\text{Src}}(T) \neq P_{\text{Tgt}}(T) \). The source and target domains exhibit significant differences in sample spaces, including shifts in feature spaces, label spaces, and conditional distributions. Cross-domain few-shot learning typically employs a multi-stage joint optimization framework, integrating meta-learning with domain adaptation techniques. Under the condition of a single visible source domain, the single-source domain data \( D_{\text{src}} = \{ (x_i, y_i) \}_{i=1}^{N_{\text{src}}} \) initializes the model parameters, ensuring the feature extractor \( f_{\theta}(\cdot) \) captures essential semantic information. The classification model comprises a feature extractor \( f_{\theta}: \mathcal{X} \rightarrow \mathbb{R}^d \) and a classifier \( h_{\phi}: \mathbb{R}^d \rightarrow \mathbb{R}^C \), where \( C \) represents the number of classes in the source domain. The decision boundaries typically converge under the constraint of cross-entropy:

\begin{equation}
\label{deqn_ex1a}
\mathcal{L}_{\text{CE}}(\theta, \phi) = -\frac{1}{N_{\text{src}}} \sum_{i=1}^{N_{\text{src}}} \sum_{c=1}^{C} y_{i,c} \log \left( \sigma\left(h_{\phi}(f_{\theta}(x_i))_c \right) \right)
\end{equation}

where \( \sigma(\cdot) \) denotes the Softmax function, and \( y_{i,c} \) represents the one-hot encoded label for class \( C \).

Jointly optimize \( \theta \) and \( \phi \) through gradient descent:

\begin{equation}
\label{deqn_theta_update}
\theta \leftarrow \theta - \eta \nabla_{\theta} \mathcal{L}_{\text{CE}}
\end{equation}

\begin{equation}
\label{deqn_phi_update}
\phi \leftarrow \phi - \eta \nabla_{\phi} \mathcal{L}_{\text{CE}}
\end{equation}

Ideally, the extracted features should satisfy cross-domain alignment \( \phi(D_s) \approx \phi(D_t) \), and maintain consistency in conditional distributions \( p_{\phi}(y \mid \phi(x)) \) across domains. However, the objectives of pre-training conflict with these cross-domain requirements:

\begin{equation}
\label{deqn_phi_update}
\min_{\phi, f} \; \mathcal{L}_{\text{CE}} 
\quad \text{vs} \quad 
\min_{\phi} \; \mathcal{R}(\phi(D_s), \phi(D_t))
\end{equation}

where \( \mathcal{R}(\cdot) \) is a regularization term measuring the discrepancy between distributions. The pre-trained features \( \phi_{\text{pre}} \) might overly fit the decision boundaries of \( D_s \), leading to distributional shifts when applied to \( D_t \).

In the single-source domain meta-training phase, the parameters \( \theta \) are further fine-tuned through task-based training:

\begin{equation}
\label{deqn_inner_update}
\theta_i' = \theta - \alpha \nabla_{\theta} \, \mathcal{L}_{S_{\text{Src}}^{\tau}}(f_{\theta}, h_{\phi})
\end{equation}

\begin{equation}
\label{deqn_outer_update}
\theta \leftarrow \theta - \beta \nabla_{\theta} \sum_{T_i} \mathcal{L}_{Q_{\text{Src}}^{\tau}}(f_{\theta_i'}, h_{\phi})
\end{equation}

The domain discrepancy can be defined as \( \Delta_{\text{domain}} = \text{KL}(p_{\phi}(D_s) \parallel p_{\phi}(D_t)) \), in meta-training, domain shifts are simulated solely by using class-disjoint tasks. Some other studies attempt to obtain a pseudo-target domain by applying data augmentation to the source domain to approximate the target distribution. However, distributional shift issues still persist:

\begin{align}
\Delta_{\text{meta-train}} &= \text{MMD}(\phi(D_s^{\text{train}}), \phi(D_s^{\text{test}})) \nonumber \\
&\ll \Delta_{\text{meta-test}} = \text{MMD}(\phi(D_s), \phi(D_t))
\end{align}

This discrepancy implies that the mapping function \( \phi_{\text{meta}} \) learned during meta-training cannot generalize effectively to the distributional shift represented by \( \Delta_{\text{meta-test}} \).

In the testing phase, rapid adaptation learning is performed on the target domain support set \( S_{\text{tgt}} = \{ (x_j, y_j) \}_{j=1}^{N \times K} \):

\begin{equation}
\label{deqn_outer_update}
\theta_{\text{tgt}}' = \theta - \alpha \nabla_{\theta} \left( \mathcal{L}_{S_{\text{tgt}}}(f_{\theta}, h_{\phi}) + \gamma \cdot \mathcal{R}(\theta) \right)
\end{equation}

where \( \mathcal{R}(\theta) \) is a regularization term introduced to address the feature consistency issue encountered during cross-domain few-shot transfer.  
Finally, performance is evaluated on the query set \( Q_{\text{tgt}} \):

\begin{small}
\begin{equation}
\label{deqn_outer_update}
\text{Accuracy} = \frac{1}{|Q_{\text{tgt}}|} \sum_{(x, y) \in Q_{\text{tgt}}} \mathbb{I} \left( \arg\max h_{\phi}(f_{\theta_{\text{tgt}}'}(x)) = y \right)
\end{equation}
\end{small}

Therefore, the existing issues in cross-domain few-shot learning at the current stage can be summarized as follows:

(1) During the pre-training stage, optimizing intra-domain maximum margin criteria through cross-entropy loss creates a conflict between the resulting feature space and the requirement for cross-domain feature matching in the meta-testing stage.

(2) When only a single source domain is used to simulate domain shifts in the meta-training stage, the generated pseudo-target domain systematically deviates from the actual distribution in real cross-domain test scenarios, causing overfitting in meta-knowledge transfer capabilities.

\begin{figure}[t]
\centerline{\includegraphics[width=\columnwidth]{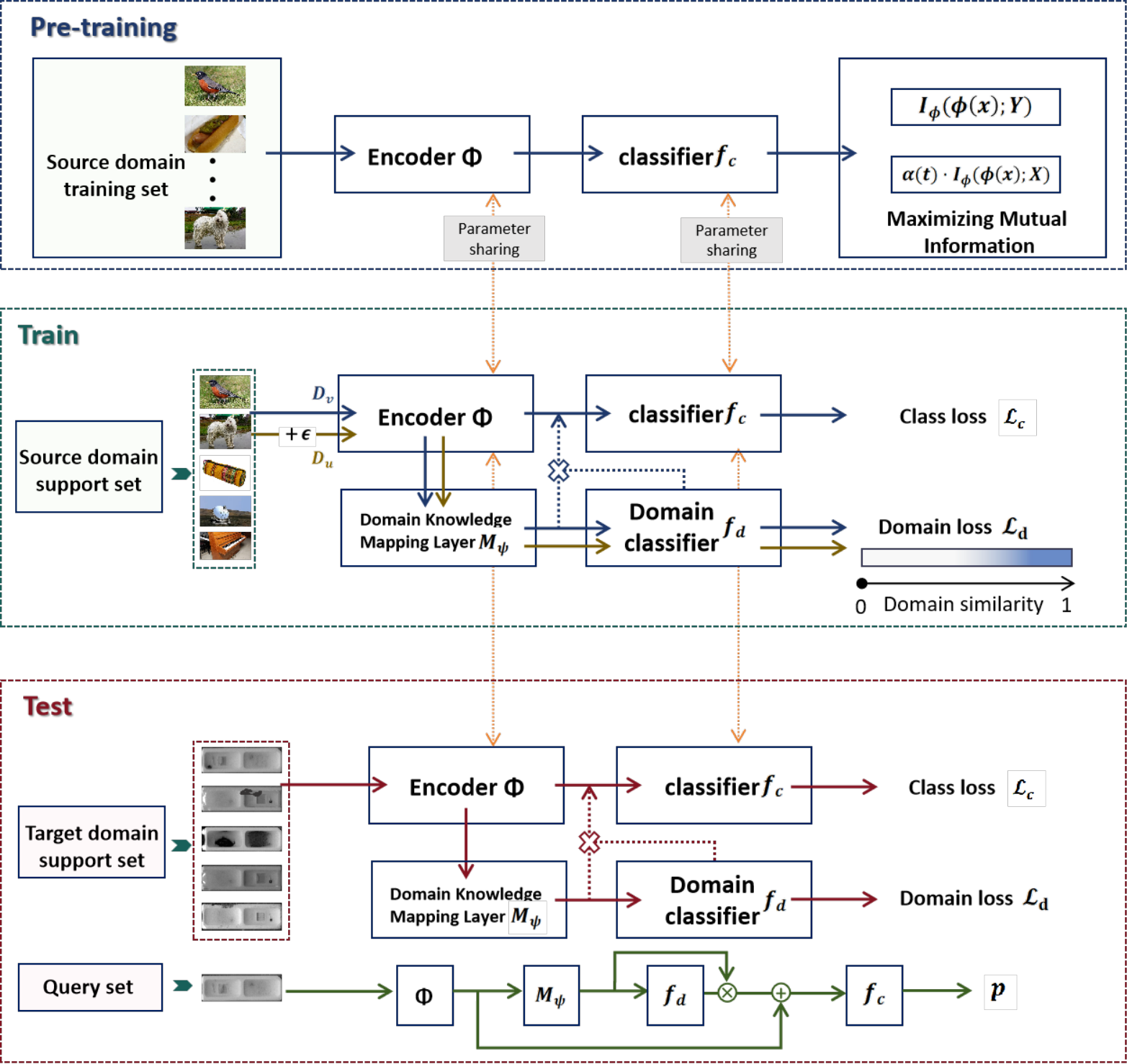}}
\caption{Framework Diagram of the Domain Knowledge Mapping-based Cross-Domain Few-Shot Learning Method.}
\label{img1}
\end{figure}

\section{A Cross-Domain Few-Shot Learning Method Based on Domain Knowledge Mapping}
The overall framework of the proposed method is shown in Figure 4.1. To address the issue of feature matching in pre-trained feature spaces, we propose a maximum mutual information-based hybrid-supervised pre-training approach. In this approach, the model is pre-trained using the source domain dataset, where features are extracted by a feature encoder and subsequently fed into a classifier to obtain prediction probabilities. The model’s loss function is constructed by maximizing mutual information, dynamically integrating self-supervised learning and supervised learning during training. This prevents mode collapse caused by relying solely on the cross-entropy loss function. 
To resolve the overfitting issue in meta-knowledge transfer, we introduce a domain-knowledge mapping method. The proposed network comprises a feature encoder, domain-knowledge mapping layer, classifier, and domain classifier. Firstly, the source domain is divided into a visible domain and a pseudo-unseen domain; the pseudo-unseen domain is generated by pixel-wise random mixing of the visible domain with Gaussian noise. Both visible and pseudo-unseen domains are passed through the feature encoder to extract features. The domain classifier assigns domain-difference labels based on the proportion of Gaussian noise mixture, inducing the domain classifier's capability to distinguish source domain similarity. Meanwhile, the domain-knowledge mapping layer gains domain transformation capabilities through gradient-reversal learning. The strength of domain knowledge mapping is dynamically adjusted based on domain similarity, enabling the model to perform adaptive domain mapping. During the meta-training tasks in the testing phase, the model fine-tunes the domain classifier and domain-knowledge mapping layer based on the target domain's features, allowing the domain classifier to better differentiate the target domain, and the domain-knowledge mapping layer to learn transferable features. Finally, in the query set of the target domain, the output gain of the knowledge mapping layer is adjusted according to domain similarity scores generated by the domain classifier, providing the model with adaptive adjustment capabilities within the query set.

\subsection{Maximal Mutual Information Mixed-Supervision Pretraining}
According to the analysis of the limitations of cross-task pre-training with cross-entropy described in Section 4.2, the cross-entropy loss can be reformulated as follows:

\begin{small}
\begin{equation}
\label{deqn_inner_update}
\mathcal{L}_{\text{CE}} = -\mathbb{E}_{(x, y) \sim D_s} \sum_{k=1}^{K} \mathbb{I}[y = k] \log \left( \frac{\exp(f_k(\phi(x)))}{\sum_{j=1}^{K} \exp(f_j(\phi(x)))} \right)
\end{equation}
\end{small}
where \( \phi(x) \) represents the pre-trained features extracted by the model, and these features \( \phi(x) \) converge under the gradient effects induced by the cross-entropy loss function.

Using only cross-entropy loss may lead to class-mode collapse \cite{142} in the semantic representation space \( \phi(x) \). Specifically, \( \phi(x) \) converges toward the mean feature representation of each class. In this case, maximizing the average information that \( Y \) obtains from \( \phi(x) \), i.e., maximizing the mutual information \( I(\phi(x); Y) \), leads to smaller conditional entropy \( H(\phi(x) \mid Y) \). This corresponds to minimal posterior uncertainty of \( Y \) given \( \phi(x) \). However, the objective of cross-domain few-shot learning is not to achieve optimal classification performance on base classes, but rather to equip \( \phi(x) \) with the capability to quickly differentiate new classes. 

Consequently, this leads to two primary issues: firstly, overfitting on base classes, where representations excessively rely on the statistical properties of base-class labels and lack comprehensive understanding of the data itself; secondly, poor generalization, as in few-shot learning scenarios, new classes may substantially differ from base-class distributions, making label-dependent representations ineffective in adapting to new classes. In contrast, the objective of self-supervision is to maximize \( I(\phi(x); X) \), the mutual information between representations and the original data. This ensures representations retain comprehensive data information rather than focusing solely on labels.

The proposed method jointly employs cross-entropy and self-supervised learning to simultaneously increase \( I(\phi(x); Y) \) and maximize \( I(\phi(x); X) \) through self-supervision. This encourages the model to learn more generalized feature representations, enabling the classifier to rapidly adapt to abrupt domain-induced feature changes during meta-training tasks.

The overall pre-training objective function is formulated as follows:

\begin{equation}
\label{deqn_inner_update}
\mathcal{E}_{\text{pre}} = \max_{\phi} \left[ I_{\phi}(\phi(x); Y) + \alpha(t) \, I_{\phi}(\phi(x); X) \right]
\end{equation}

where \( \alpha(t) \) is a dynamic weight that increases linearly with the training iteration \( t \), defined as \( \alpha(t) = \kappa \cdot \frac{t}{T} \). Thus, in the early stage of pre-training, supervised learning dominates, while self-supervised and cross-domain constraints gradually strengthen in later stages, alleviating the mismatch between pre-training and cross-domain requirements.

Maximizing only \( I_{\phi}(\phi(x); Y) \) leads to an increase in \( I_{\phi}(\phi(x) \mid X) \), resulting in feature discarding of task-irrelevant information, thus impairing cross-domain generalization. Conversely, maximizing \( I_{\phi}(\phi(x); X) \) is equivalent to minimizing \( I_{\phi}(\phi(x) \mid X) \), forcing \( Z \) to retain more information from the original data and consequently improving cross-domain adaptability. Since \( I(\phi(x); X) = I(\phi(x); Y) + I(\phi(x); X \mid Y) \), maximizing \( I(\phi(x); X) \) not only preserves label-related information but also captures label-independent features crucial for cross-domain tasks, alleviating source-domain overfitting.

Through self-supervised learning, \( I_{\phi}(\phi(x); X) \) can be expressed as follows:
\begin{small}
\begin{align}
\label{deqn_inner_update}
I_{\phi}(\phi(x); X) = \mathbb{E}_{x \sim D_s} \Bigg[ 
&\frac{\phi(x^1)^\top \phi(x^2)}{\tau} \nonumber \\
&- \log \sum_{x_k \in B} \exp \left( \frac{\phi(x^1)^\top \phi(x_k)}{\tau} \right) 
\Bigg]
\end{align}
\end{small}

In the construction of positive and negative samples during self-supervision, previous methods typically rely solely on self-augmentation, neglecting class distinctions. To enhance adaptation to novel class variations during pre-training, new classes are introduced explicitly as negative samples in the self-supervised formulation, while expressions from the same class are excluded.


\begin{small}
\begin{align}
\label{deqn_inner_update}
I_{\phi}(\phi(x); X) &= \mathbb{E}_{x \sim S\cup U} \Bigg[
\frac{\phi(x^1)^\top \phi(x^2)}{\tau} \nonumber\\
&- \log \sum_{x_k \in B_S \cup B_Q} \exp \left( \frac{\phi(x^1)^\top \phi(x_k)}{\tau} \right)\nonumber\\ 
&- \sum_{x_m \in (Y|X)} \exp  (\frac{\phi(x^1)^\top \phi(x_k)}{\tau})
\Bigg]
\end{align}
\end{small}


\subsection{Domain Knowledge Mapping}
To learn the distribution patterns across domains from meta-training to meta-testing, this section proposes a domain knowledge mapping approach. By assessing the transfer difficulty between visible and unseen domains, the method dynamically adjusts the weights of domain knowledge mapping to accommodate tasks of varying migration difficulty. Domain knowledge mapping operates during the meta-training process and comprises a feature encoder \( \phi \), a domain knowledge mapping layer \( \mathcal{M}_{\psi} \), a classifier \( f_c \), and a domain classifier \( f_d \). The data domains include a visible domain \( D_v \) and a pseudo-unseen domain \( D_u \). The visible domain \( D_v \) is directly sampled from the source domain support set \( S_{\text{src}} \), while the pseudo-unseen domain \( D_u \) is sampled from the source domain support set as follows:

\begin{equation}
\label{deqn_inner_update}
x_u = \lambda x_v + (1 - \lambda) \, \epsilon, \quad \epsilon \sim \mathcal{N}(\mu_{D_v}, \sigma_{D_v}^2)
\end{equation}

where \( \lambda \) is the mixing ratio, with higher proportions of Gaussian noise indicating increased transfer difficulty. \( \mu_{D_v} \) and \( \sigma_{D_v} \) represent the pixel-wise mean and standard deviation of the visible domain, respectively.

First, support-set tasks \( T_i^S = \{ X_v^S, X_u^S, Y^S \} \) are sampled from the visible and pseudo-unseen domains, and then provided as inputs to the feature encoder:

\begin{equation}
\label{deqn_inner_update}
z_{v/u} = \phi(x_{v/u})
\end{equation}

The transfer difficulty between visible and unseen domains is measured through the learning of a discriminator. The discriminator's loss function is as follows:
\begin{small}
\begin{equation}
\label{deqn_inner_update}
\mathcal{L}_d(D_v^S, D_u^S) = \max_{f_d} \left[ \mathbb{E}_{x_v \in D_v^S} \log(\rho_v) + \mathbb{E}_{x_u \in D_u^S} \log(d_u - \rho_v) \right]
\end{equation}
\end{small}

where \( d_u = 1 - \lambda \) is the pseudo-unseen domain label, and \( \lambda \) denotes the mixing ratio. Additionally, \( \rho_v = f_d(\mathcal{M}_{\psi}(z_v)) \) represents the domain transfer difficulty score. By adjusting the mixing ratio \( \lambda \), the migration difficulty of pseudo-unseen domain samples is simulated, enabling the discriminator to learn to differentiate samples based on domain transfer difficulty.

The generator guides the pseudo-unseen domain to align with the visible domain:

\begin{equation}
\label{deqn_inner_update}
\mathcal{L}_g(D_v^S, D_u^S) = \min_{\phi} \, \mathbb{E}_{x_u \in D_u^S} \log \left( d_u - f_d(\mathcal{M}_{\psi}(z_u)) \right)
\end{equation}

The domain knowledge mapping layer learns mapping transformations via the pseudo-target domain and integrates these transformations with encoder features. This enables the encoded vectors to carry domain-specific knowledge, dynamically adapting to variations in domain transfer difficulty through weighting by the transfer difficulty score. The input features for the classifier can be formulated as follows:

\begin{equation}
\label{deqn_inner_update}
c_v = z_v + \rho_v \, \mathcal{M}_{\psi}(z_v)
\end{equation}

The meta-model's initialized parameters are adapted using the support set, while the model parameters are updated only on the query set. The parameter update rules are as follows:

\begin{equation}
\label{deqn_update_phi}
\theta_{\phi} \leftarrow \theta - \eta \nabla \mathcal{L}_c
\end{equation}

\begin{equation}
\label{deqn_update_f}
\theta_f \leftarrow \theta - \eta \nabla \mathcal{L}_{\text{cls}} + \eta \nabla \mathcal{L}_d
\end{equation}

\begin{equation}
\label{deqn_update_m}
\theta_{\mathcal{M}_{\psi}} \leftarrow \theta - \eta \nabla \mathcal{L}_g
\end{equation}

where \( \mathcal{L}_{\text{cls}} \) is the classification loss based on cross-entropy, computed between predictions \( (c_v, y) \).

In the meta-testing phase, set \( d_u = 1 \) in \( \mathcal{L}_d \), and perform a single calibration of the domain knowledge mapping layer using the support set. Then, dynamically adjust the features in the query set through the domain knowledge mapping layer based on the sample transfer difficulty:

\begin{equation}
\label{deqn_update_phi}
c_{\text{tgt}} = z_{\text{tgt}} + \rho_{\text{tgt}} \, \mathcal{M}_{\psi}(z_{\text{tgt}}) 
\quad \Rightarrow \quad 
p_t = f_d(c_{\text{tgt}})
\end{equation}

\section{Experiment}
\subsection{Dataset Introduction}
In this experiment, mini-ImageNet is utilized as the source domain dataset, comprising 100 classes with 600 images per class. The training set for the source domain randomly selects 64 classes from mini-ImageNet as base classes, while the remaining 16 classes form the source domain test set to evaluate the generalization ability of the pretrained model. The target domain datasets include CUB, Cars, Places, Plantae, and an LED packaging defect detection dataset. The LED dataset encompasses binary classification (defect detection) and multi-class defect classification tasks, facilitating evaluation of cross-domain few-shot learning performance. Figure 4.2 illustrates the difficulty levels associated with few-shot classification tasks across these datasets and quantifies the degree of domain shift. The domain shift is measured using the Earth Mover’s Distance (EMD), while the difficulty level of the few-shot classification tasks is derived from the performance obtained in few-shot experiments.

\textbf{CUB dataset} contains 200 bird species with a total of 11,788 images. Each image is annotated with extensive fine-grained attributes, including beak shapes and feather colors. It is characterized by small inter-class differences and high feature similarity among samples, leading to high transfer difficulty. The CUB dataset is frequently employed to evaluate model performance on fine-grained classification tasks, specifically assessing the model’s capability to capture subtle feature differences.

\textbf{Cars dataset} comprises 196 automobile classes with a total of 16,185 images, covering various car brands, models, and production years. The dataset features considerable inter-class differences but complex image backgrounds, including streets and showrooms, which results in moderate transfer difficulty. It is particularly suitable for evaluating the model's feature extraction capability in complex background scenarios, as well as adaptability to significant inter-class variations.

\textbf{Places dataset} contains 365 scene categories with approximately 1.8 million images, including natural landscapes, urban architectures, and indoor environments. Its primary characteristics are pronounced inter-class differences and diverse sample features, resulting in relatively low transfer difficulty. This dataset is effective for assessing a model’s generalization capability in large-scale scene classification tasks and robustness to diverse features.

\textbf{Plantae dataset} includes 100 plant species with a total of 10,000 images, covering various leaves, flowers, and fruits. Its distinguishing features are small inter-class differences and complex sample attributes, such as leaf textures and flower shapes, leading to higher transfer difficulty. It is appropriate for evaluating model performance on fine-grained plant classification tasks and its ability to extract complex texture features.

\textbf{LED dataset} consists of two classification tasks: binary classification (identifying whether an LED package is defective or not) and multi-class classification (identifying specific types of LED package defects). The dataset's features are primarily homogeneous and mainly focused on fine-grained texture distinctions. While the binary task is relatively easy due to its clear objective, the multi-class classification involves various intricate defect categories, thereby posing greater difficulty. The LED dataset is ideal for assessing the practicality of models in industrial inspection tasks and their capability to rapidly adapt to simple features.

Overall, these datasets significantly differ in terms of category distribution, sample characteristics, and transfer difficulty. Collectively, they enable a comprehensive evaluation of the generalization capability and adaptability of cross-domain few-shot learning methods.

\begin{figure}[t]
\centerline{\includegraphics[width=\columnwidth]{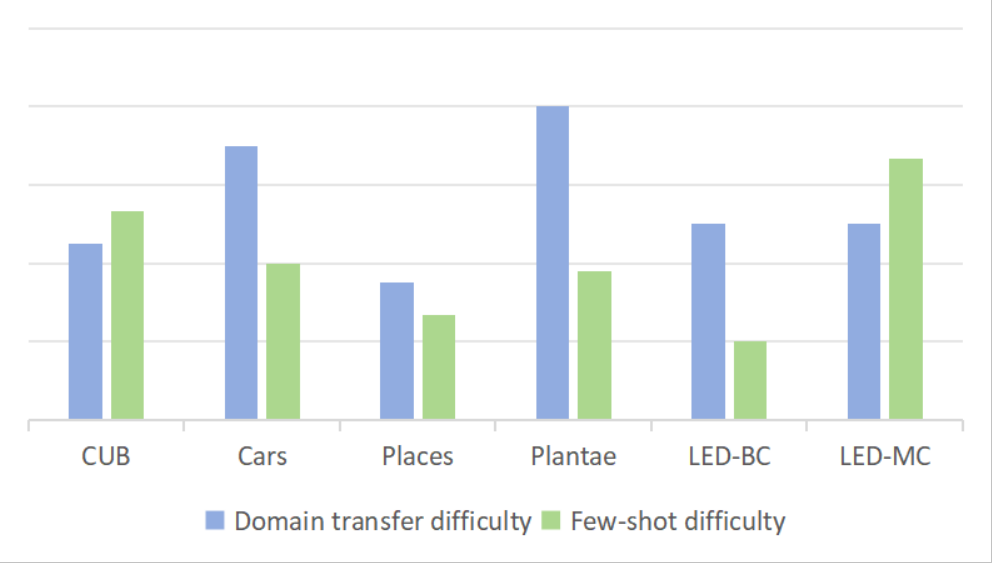}}
\caption{Relative relationship between cross-domain difficulty and small sample difficulty in different dataset domains.}
\label{img2}
\end{figure}

The target domain tasks are divided into 5-way 1-shot and 5-way 5-shot configurations, meaning each task comprises five novel classes, with only one or five support samples provided per class. Specifically, for the LED binary classification task, a 2-way 1-shot and 2-way 5-shot configuration is employed. All target domain datasets are excluded from both the pre-training and meta-training stages, being used exclusively for cross-domain evaluation during the meta-testing stage to ensure genuine distributional differences between domains. Data augmentation techniques applied include random cropping, horizontal flipping, and normalization.

\subsection{Experiment Setup}
The proposed method is implemented based on the PyTorch 1.9.0 framework, with experiments conducted on a server equipped with a single GeForce RTX 4090 GPU. The proposed approach incorporates a maximum mutual information mixed-supervision strategy during the pre-training phase, jointly optimizing cross-entropy loss and self-supervised loss. The feature encoder \( \phi \) employs ResNet-10 as the backbone network, producing feature representations with an output dimension of 512. The self-supervised branch leverages contrastive learning to maximize mutual information between encoded features and the original data. The contribution balance between supervised and self-supervised learning is regulated by a dynamic weighting factor \( \alpha(t) \), which increases linearly with training iterations, starting from an initial value \( \kappa = 0.1 \), and reaching its maximum at the total number of training epochs \( T \). 

The domain knowledge mapping layer \( \mathcal{M}_{\psi} \) consists of a two-layer fully connected neural network, which takes encoded features \( z \) as input and outputs domain-transfer weights \( \rho_v \). During the meta-training phase, the Adam optimizer is adopted with an initial learning rate of \( 1\text{e}{-4} \), batch size of 32, and training duration of 200 epochs. The domain knowledge mapping layer generates pseudo-unseen-domain samples via a mixing ratio \( \lambda \in [0.3, 0.7] \).

In the meta-testing phase, 1{,}000 testing tasks are randomly generated for each dataset to compute the average accuracy. In each testing task, five categories are randomly selected, with \( N_S \) images per category randomly chosen as the support set for updating meta-network parameters, followed by \( N_Q \) images per category selected as the query set. The final reported results are the average accuracies across these 1{,}000 query sets.

\subsection{Results and Analysis}
Three mainstream meta-learning networks are adopted as baseline models: MatchingNet \cite{a143}, RelationNe \cite{a144}, and GNN \cite{a145}. Specifically, MatchingNet employs an attention-based mechanism to compute cosine similarities between query and support samples. RelationNet introduces a learnable relation module to measure pairwise sample similarities and outputs similarity scores. The GNN method utilizes graph neural networks to model relationships among samples within a task, propagating label information through graph structures to optimize classification decisions. Additionally, comparative methods include traditional meta-training (-), the Feature Transformation layer (FT) \cite{a138}, and Adversarial Task Augmentation (ATA) \cite{a146}. FT enhances image features using a parameterized feature transformation layer during meta-testing, whereas ATA improves the model's adaptability to novel categories by augmenting tasks adversarially.

Tables 4.1 and 4.2 present the classification results for cross-domain few-shot learning under 1-shot and 5-shot meta-learning settings, respectively, with the best results highlighted in bold. The experimental results in Tables 4.1 and 4.2 demonstrate that the proposed method (OUR) consistently outperforms traditional meta-learning methods (MatchingNet, RelationNet, and GNN), as well as other cross-domain few-shot learning methods (FT and ATA), across both 5-way/2-way 1-shot and 5-shot settings. Experimental validation indicates that the core challenge of cross-domain few-shot learning—learning generalized representations from the source domain (mini-ImageNet) for rapid adaptation to target domain distribution discrepancies—is effectively addressed by our approach. Specifically, our dual-optimization strategy, integrating maximum mutual information mixed-supervision pretraining and domain-knowledge mapping, substantially mitigates cross-task and cross-domain mismatch issues.

\begin{table*}[!ht]
    \caption{{5-way/2-way 1-shot setting the classification result of cross-domain few-shot classification (\%)}}
\label{table_2}
\centering
\resizebox{1.7\columnwidth}{!}{
    \begin{tabular}{ccccccccc}
    \hline\hline
        Few-shot & Cross-domain & CUB & Cars & Places & Plante & LED(MC) & LED(BC) \\ 
        method & method & ~ & 5-way 1-shot & ~ & ~ & ~ & 2-way 1-shot \\ \hline
        ~ & - & 36.12 & 31.05 & 50.12 & 32.91 & 52.71 & 87.54  \\ 
        MatchingNet & FT & 36.88 & 30.12 & 51.32 & 34.72 & 53.91 & 88.42\\ 
        ~ & ATA & 39.93 & 32.46 & 53.87 & 36.59 & 55.84 & 90.21  \\ 
        ~ & OURS & 41.23 & 33.69 & 54.9 & 37.78 & 57.52 & 90.32 \\ 
        ~ & - & 44.72 & 31.9 & 52.68 & 33.78 & 53.36 & 89.31 \\ 
        RelationNet& FT & 45.74 & 32.43 & 53.7 & 32.73 & 53.78 & 88.06 \\ 
        ~ & ATA & 45.24 & 33.8 & 54.3 & 34.59 & 57.64 & 90.03 \\ 
        ~ & OURS & 47.1 & 34.42 & 53.84 & 36.95 & 58.81 & 90.61\\ 
        ~ & - & 48.56 & 32.58 & 56.44 & 37.58 & 58.94 & 89.98 \\ 
        GNN & FT & 46.47 & 30.63 & 52.71 & 32.62 & 56.36 & 89.71\\ 
        ~ & ATA & 50.51 & 34.35 & 57.31 & 40.03 & 60.43 & 90.42 \\ 
        ~ & OURS & 51.09 & 38.61 & 60.57 & 40.47 & 60.24 & 90.87\\ \hline
    \end{tabular}}
\end{table*}

In the 5-way 1-shot scenario presented in Table~4.1, the proposed method exhibits particularly strong performance on datasets with high transfer difficulty, such as CUB and Plantae. Specifically, in the CUB task using GNN as the baseline, the traditional meta-training (--) achieves an accuracy of 48.56\%, while our method (OUR) attains 51.09\%, outperforming the second-best method ATA (50.51\%) by 0.58 percentage points. This improvement is primarily attributed to the mixed-supervision pretraining, which significantly enhances fine-grained feature extraction. Fine-grained bird classification in the CUB dataset relies on subtle local attributes such as feather textures and beak shapes. Traditional cross-entropy pretraining tends to overly depend on class centroids, causing category-mode collapse. 

Conversely, our self-supervised branch maximizes the mutual information \( I(\phi(x); X) \), forcing the model to preserve more label-agnostic yet cross-domain critical detailed features. Incorporating novel-class samples as negative pairs during contrastive self-supervised learning prevents excessive clustering of same-class samples, thus improving the model's ability to distinguish unseen categories. Additionally, the accuracy of 40.47\% on the Plantae dataset further validates our method’s effectiveness in large-scale fine-grained scenario classification tasks. The introduction of a dynamic weighting factor \( \alpha(t) \) allows supervised learning to dominate in the early stages of pretraining, gradually increasing self-supervised constraints in later stages. This dynamic balancing enables the model to simultaneously capture global shapes and local textures, alleviating the suppression of complex features caused by single-source supervision.

For the Cars dataset, characterized by moderate transfer difficulty, our method achieves an accuracy of 38.61\% under the 5-way 1-shot setting with the GNN framework, representing a substantial improvement of 6.03\% over the traditional meta-training (--) baseline, and notably surpassing FT and ATA methods. The main challenges of the Cars dataset lie in complex background interference such as exhibition lighting and occlusions, as well as subtle differences between similar car models. 

Our domain-knowledge mapping layer dynamically adjusts feature fusion weights \( \rho_v \) by simulating the transfer difficulty from pseudo-unseen domains \( D_u \), thus enabling the model to distinguish between task-relevant features and domain-related noise. The visible domain features \( z_v \) are fused with the mapping output \( \mathcal{M}_{\psi}(z_v) \), forming \( c_v = z_v + \rho_v \cdot \mathcal{M}_{\psi}(z_v) \), thereby improving the robustness of the model against background variations.

Furthermore, performance significantly improves under the 5-shot scenario, with the accuracy on the Cars dataset increasing nearly 10 percentage points compared to the 1-shot scenario. This demonstrates that a larger number of support samples aids in more accurately estimating transfer difficulty weights, ultimately optimizing the feature fusion process.

\begin{table*}[!ht]
    \caption{{5-way/2-way 5-shot setting the classification result of cross-domain few-shot classification (\%)}}
\label{table_3}
\centering
\resizebox{1.7\columnwidth}{!}{
    \begin{tabular}{ccccccccc}
    \hline\hline
        Few-shot & Cross-domain & CUB & Cars & Places & Plante & LED(MC) & LED(BC)  \\ 
        method & method & ~ & 5-way 5-shot & ~ & ~ & ~ & 2-way 5-shot  \\ \hline
        ~ & - & 52.40 & 39.82 & 64.42 & 47.42 & 80.91 & 91.54  \\ 
        MatchingNet & FT & 55.74 & 41.55 & 65.30 & 42.52 & 79.82 & 92.37  \\ 
        ~ & ATA & 57.68 & 46.01 & 68.21 & 51.28 & 82.34 & 90.21  \\ 
        ~ & OURS & 59.51 & 46.38 & 69.05 & 52.61 & 85.92 & 94.18  \\ 
        ~ & - & 63.13 & 43.90 & 71.23 & 48.74 & 81.56 & 92.82  \\ 
        RelationNet & FT & 65.30 & 46.42 & 70.96 & 49.92 & 82.67 & 92.34  \\ 
        ~ & ATA & 68.57 & 49.40 & 75.83 & 52.95 & 83.78 & 93.76  \\ 
        ~ & OURS & 66.54 & 49.52 & 76.57 & 54.52 & 84.23 & 93.67  \\ 
        ~ & - & 63.78 & 44.76 & 71.62 & 51.18 & 84.45 & 90.75  \\ 
        GNN & FT & 58.44 & 34.18 & 67.06 & 43.40 & 83.33 & 90.45  \\ 
        ~ & ATA & 65.59 & 47.18 & 72.43 & 55.33 & 85.12 & 95.58  \\ 
        ~ & OURS & 66.12 & 48.12 & 73.06 & 55.91 & 86.07 & 96.47 \\ \hline
    \end{tabular}}
\end{table*}

The advantages of the proposed method are similarly pronounced in low-transfer-difficulty tasks such as Places and LED. On the diverse Places dataset, our method achieves an accuracy of 60.57\% under the 5-way 1-shot scenario (using the GNN framework), outperforming the baseline (--, 56.44\%) by 4.13 percentage points. Furthermore, the accuracy improves to 73.06\% with increased support samples (5-shot scenario). These results validate the inclusiveness of the maximum mutual information pretraining toward diverse features; the self-supervised branch leverages contrastive learning to encourage the model to capture global layouts and local semantics of scenes rather than relying solely on a single discriminative region.

In the LED multi-class task under the 5-way 1-shot setting (GNN framework), our method achieves an accuracy of 60.24\%, slightly lower (by 0.19 percentage points) than ATA (60.43\%). However, in the 5-shot scenario, our method outperforms ATA (OUR: 86.07\% vs. ATA: 85.12\%). This phenomenon may be attributed to the interference of self-supervised noise on limited support samples under the 1-shot condition. Conversely, the increased support samples in the 5-shot setting mitigate this issue, while the domain-knowledge mapping layer enhances discrimination efficiency for simple features such as brightness and color by calibrating transfer weights \( \rho_{\text{tgt}} \).

The proposed method demonstrates superior performance in the LED binary classification task. Specifically, under the GNN framework, the defect classification accuracy reaches 90.87\% in the 5-shot scenario. Compared to manual visual inspection, which achieves an accuracy of approximately 95\%, this result indicates that using just five defect samples meets the practical demands of LED package defect detection, enabling rapid deployment with minimal samples. The simplicity of LED binary classification (normal/defective) facilitates faster model convergence, and the dynamic weighting mechanism of domain-knowledge mapping further amplifies cross-domain adaptability. In the pretraining phase, the mixed-supervision strategy enables the model to initially capture common defect features. During meta-testing, the pseudo-unseen domain noise simulates real industrial conditions such as variations in lighting and defect characteristics, prompting the model to quickly filter domain-related interference and focus on the intrinsic task-related features.

The Feature Transformation (FT) layer exhibits unstable performance across multiple tasks. For instance, under the GNN framework in the Cars dataset (5-way 1-shot), FT achieves an accuracy of 30.63\%, lower than the original meta-training (32.58\%), indicating that the manually parameterized feature transformation is prone to overfitting due to the limited target-domain samples. ATA improves model generalization through task-level data augmentation, but it relies on fixed augmentation strategies, limiting its adaptability to abrupt inter-domain distribution shifts. In contrast, our method synergistically optimizes self-supervision and domain-knowledge mapping, achieving dynamic adaptation from pretraining to meta-training. Specifically, self-supervision enhances feature invariance across tasks

\subsection{Ablation Experiment}
To validate the contribution of each component within the maximum mutual information mixed-supervision pretraining strategy, three pretraining configurations are compared: 

(1) \textbf{Supervised Learning:} pretraining solely with cross-entropy loss;  

(2) \textbf{Self-Supervised Learning:} pretraining by maximizing mutual information \( I(\phi(x); X) \) through contrastive learning only; 

(3) \textbf{Maximum Mutual Information Mixed-Supervision:} jointly optimizing cross-entropy loss and self-supervised loss, introducing a dynamic weighting factor \( \alpha(t) \) to balance their contributions.

The experiments are conducted under a 5-way 5-shot setting, covering target domains including CUB, Cars, Places, Plantae, and LED datasets, with the feature encoder employing the GNN framework. All other hyperparameters remain consistent with the main experiments.

\begin{table*}[!ht]
    \caption{The cross-domain few-shot classification results of 5-way 5-shot under different supervised pre-training settings (\%)}
\label{table_3}
\centering
\resizebox{1.7\columnwidth}{!}{
    \begin{tabular}{ccccccccc}
    \hline\hline
5-way 5-shot & CUB & Cars & Places & Plante & LED(MC) & LED(BC)  \\ \hline
        Supervised Learning & 68.16 & 48.12 & 71.98 & 54.65 & 84.87 & 92.54  \\ 
        Self-Supervised Learning & 49.43 & 43.75 & 70.64 & 49.14 & 80.16 & 84.52  \\ 
        Maximum Mutual Information Mixed Supervision & 66.12 & 47.32 & 73.06 & 55.91 & 86.07 & 96.47 \\ \hline
    \end{tabular}}
\end{table*}

From the results in Table~4.3, it can be observed that, consistent with most existing studies, purely supervised learning demonstrates robust performance on the majority of datasets, particularly excelling in tasks with high transfer difficulty, such as CUB (68.16\%) and Cars (54.65\%). This phenomenon aligns with expectations, as supervised learning directly optimizes classification boundaries using labels, effectively capturing fine-grained discriminative features. However, its performance on the LED binary classification task is inferior to the mixed-supervision method, indicating that pure supervised learning may compromise cross-domain generalization by overfitting base-class labels in simpler tasks. In the LED binary classification, defect features lack direct corresponding categories in the source domain (mini-ImageNet); thus, the supervised model excessively relies on brightness statistics learned during pretraining, whereas the mixed-supervision method preserves raw data information through the self-supervised branch, enabling better adaptation to target domain discrepancies.

Purely self-supervised learning exhibits considerable performance fluctuations. For example, in the CUB task, its accuracy declines by 18.73\% compared to supervised learning, highlighting the limitations of using solely self-supervision for capturing fine-grained features. Contrastive learning constructs positive and negative pairs through data augmentation, but the critical local details in bird images are easily disrupted by cropping and flipping, resulting in insufficient differentiation between highly similar classes. However, in the Places scene classification task, self-supervised learning demonstrates a smaller performance gap compared to supervised learning, suggesting that global semantics of scenes are robust to augmentation operations, and maximizing mutual information \( I(\phi(x); X) \) effectively models diverse features. It is noteworthy that the lower performance of self-supervised learning in the Plantae dataset compared to supervised learning further confirms fine-grained tasks' strong reliance on label supervision.

The maximum mutual information mixed-supervision method achieves optimal or near-optimal performance in most tasks. On the CUB dataset, its accuracy is 2.04\% lower than pure supervised learning but significantly higher than pure self-supervised learning. This indicates that although additional information introduced by the self-supervised branch might introduce minor noise, it overall strengthens cross-domain adaptability. Specifically, in fine-grained CUB classification, the self-supervised branch forces the model to attend to local invariances in feather texture, while the supervised branch ensures inter-class discriminability. The dynamic balance between these branches mitigates the risk of overfitting inherent in purely supervised learning.

This experiment validates the effectiveness of maximum mutual information mixed-supervision pretraining, which enhances cross-domain generalization capability while maintaining discriminative features through joint supervised and self-supervised optimization. The introduction of the dynamic weighting factor \( \alpha(t) \) balances optimization goals across training stages, effectively overcoming inherent limitations associated with purely supervised or self-supervised learning. Despite occasional fluctuations in performance, the mixed-supervision method exhibits substantial comprehensive advantages for cross-domain few-shot learning, offering a promising optimization direction for model pretraining in complex scenarios.

\subsection{Parameter Sensitivity Analysis}
\begin{figure}[t]
\centerline{\includegraphics[width=\columnwidth]{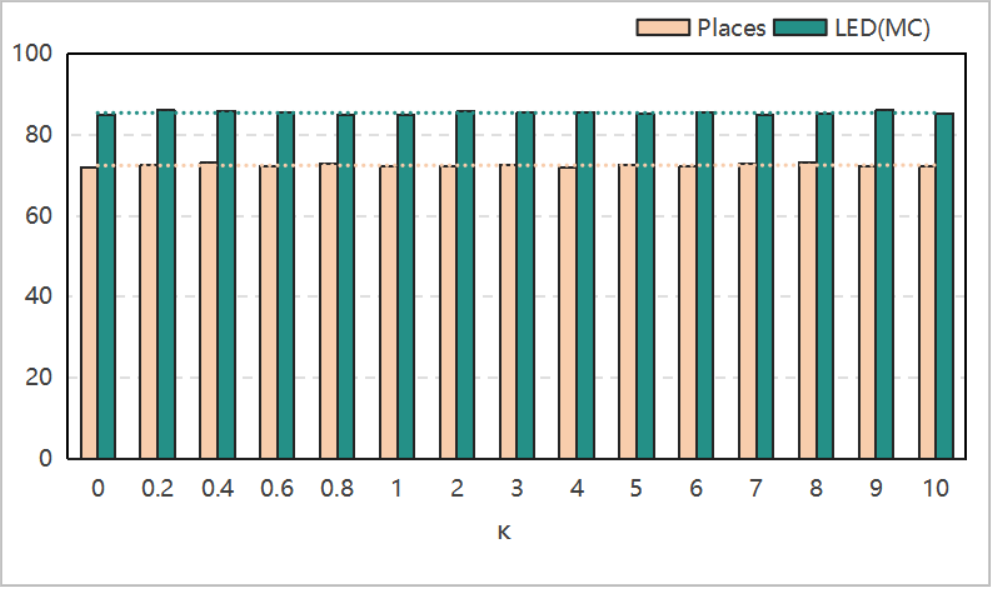}}
\caption{The results of different dynamic weight parameters in the maximum mutual information semi-supervised pre-training on the Places and LED (multi-classification) datasets.}
\label{img3}
\end{figure}

To investigate the effect of the dynamic weighting coefficient \( \kappa \) on model performance, experiments were conducted using a dynamic weighting function defined as \( \alpha(t) = \kappa t / T \), where \( T \) represents the total number of training epochs and \( \kappa \) is an adjustable scaling factor. The tested \( \kappa \) values ranged from 0 to 10, evaluated at intervals of 0.2 \((\kappa \in \{0, 0.2, 0.4, \ldots, 10\})\). The classification accuracy under a 5-way 5-shot setting was assessed on the Places and LED (multi-class) datasets, with all other hyperparameters kept constant. 

Figure~4.3 demonstrates the variation in accuracy as \( \kappa \) changes. For the Places dataset, accuracy exhibits a bimodal distribution, peaking at \( \kappa = 0.4 \) and \( \kappa = 8 \). In contrast, accuracy on the LED dataset fluctuates within a narrow range (less than 1\%) for \( \kappa \) values between 0.2 and 9. These observations validate the hypothesis that complex tasks require finely tuned dynamic weighting mechanisms, whereas structured tasks allow for broader parameter constraints. Additionally, the experiments indicate performance fluctuations for both datasets when \( \kappa > 1 \), suggesting that excessively large self-supervised weights may cause the optimization direction to deviate from the target domain.

\section{Conclusion}
we extend the cross-domain few-shot learning assumption explored, specifically addressing scenarios where the category intersection between domains is empty. Aiming at the problem of meta-learning's performance degradation under domain shifts, solutions are proposed from three dimensions. Firstly, during the pretraining phase, mixed-supervision training based on maximum mutual information is employed. Experimental results validate that the mixed-supervision approach, by retaining original data information through the self-supervised branch, effectively enhances adaptability to target domain discrepancies. Secondly, the domain-knowledge mapping mechanism is applied adaptively during both the training and testing stages, dynamically adjusting the strength of domain-knowledge mapping to better accommodate varying degrees of transfer difficulty. The experimental results confirm the robustness and effectiveness of the proposed approach.

\bibliographystyle{IEEEtran}
\bibliography{references}

\begin{thebibliography}{10}
\providecommand{\url}[1]{#1}
\csname url@samestyle\endcsname
\providecommand{\newblock}{\relax}
\providecommand{\bibinfo}[2]{#2}
\providecommand{\BIBentrySTDinterwordspacing}{\spaceskip=0pt\relax}
\providecommand{\BIBentryALTinterwordstretchfactor}{4}
\providecommand{\BIBentryALTinterwordspacing}{\spaceskip=\fontdimen2\font plus
\BIBentryALTinterwordstretchfactor\fontdimen3\font minus \fontdimen4\font\relax}
\providecommand{\BIBforeignlanguage}[2]{{%
\expandafter\ifx\csname l@#1\endcsname\relax
\typeout{** WARNING: IEEEtran.bst: No hyphenation pattern has been}%
\typeout{** loaded for the language `#1'. Using the pattern for}%
\typeout{** the default language instead.}%
\else
\language=\csname l@#1\endcsname
\fi
#2}}
\providecommand{\BIBdecl}{\relax}
\BIBdecl

\bibitem{a120}
C.~Finn, P.~Abbeel, and S.~Levine, ``Model-agnostic meta-learning for fast adaptation of deep networks,'' in \emph{International conference on machine learning}.\hskip 1em plus 0.5em minus 0.4em\relax PMLR, 2017, pp. 1126--1135.

\bibitem{a114}
J.~Snell, K.~Swersky, and R.~Zemel, ``Prototypical networks for few-shot learning,'' \emph{Advances in neural information processing systems}, vol.~30, 2017.

\bibitem{a138}
H.-Y. Tseng, H.-Y. Lee, J.-B. Huang, and M.-H. Yang, ``Cross-domain few-shot classification via learned feature-wise transformation,'' \emph{arXiv preprint arXiv:2001.08735}, 2020.

\bibitem{a79}
Y.~Hu and A.~J. Ma, ``Adversarial feature augmentation for cross-domain few-shot classification,'' in \emph{European conference on computer vision}.\hskip 1em plus 0.5em minus 0.4em\relax Springer, 2022, pp. 20--37.

\bibitem{a139}
P.~Li, F.~Liu, L.~Jiao, S.~Li, L.~Li, X.~Liu, and X.~Huang, ``Knowledge transduction for cross-domain few-shot learning,'' \emph{Pattern Recognition}, vol. 141, p. 109652, 2023.

\bibitem{a140}
J.~Oh, S.~Kim, N.~Ho, J.-H. Kim, H.~Song, and S.-Y. Yun, ``Understanding cross-domain few-shot learning based on domain similarity and few-shot difficulty,'' \emph{Advances in Neural Information Processing Systems}, vol.~35, pp. 2622--2636, 2022.

\bibitem{a141}
M.~Yazdanpanah, A.~A. Rahman, M.~Chaudhary, C.~Desrosiers, M.~Havaei, E.~Belilovsky, and S.~E. Kahou, ``Revisiting learnable affines for batch norm in few-shot transfer learning,'' in \emph{Proceedings of the IEEE/CVF conference on computer vision and pattern recognition}, 2022, pp. 9109--9118.

\bibitem{a121}
S.~Ravi and H.~Larochelle, ``Optimization as a model for few-shot learning,'' in \emph{International conference on learning representations}, 2017.

\bibitem{a69}
F.~Yao, ``Cross-domain few-shot learning with unlabelled data,'' \emph{arXiv preprint arXiv:2101.07899}, 2021.

\bibitem{a70}
C.~P. Phoo and B.~Hariharan, ``Self-training for few-shot transfer across extreme task differences,'' \emph{arXiv preprint arXiv:2010.07734}, 2020.

\bibitem{a71}
S.~Samarasinghe, M.~N. Rizve, N.~Kardan, and M.~Shah, ``Cdfsl-v: Cross-domain few-shot learning for videos,'' in \emph{Proceedings of the IEEE/CVF international conference on computer vision}, 2023, pp. 11\,643--11\,652.

\bibitem{a72}
T.~Adler, J.~Brandstetter, M.~Widrich, A.~Mayr, D.~Kreil, M.~Kopp, G.~Klambauer, and S.~Hochreiter, ``Cross-domain few-shot learning by representation fusion,'' \emph{arXiv preprint arXiv:2010.06498}, 2020.

\bibitem{a73}
Y.~Zou, S.~Zhang, J.~Yu, Y.~Tian, and J.~M. Moura, ``Revisiting mid-level patterns for cross-domain few-shot recognition,'' in \emph{Proceedings of the 29th ACM International Conference on Multimedia}, 2021, pp. 741--749.

\bibitem{a74}
M.~Li, R.~Wang, J.~Yang, L.~Xue, and M.~Hu, ``Multi-domain few-shot image recognition with knowledge transfer,'' \emph{Neurocomputing}, vol. 442, pp. 64--72, 2021.

\bibitem{a75}
L.~Yalan and W.~Jijie, ``Cross-domain few-shot classification through diversified feature transformation layers,'' in \emph{2021 IEEE International Conference on Artificial Intelligence and Computer Applications (ICAICA)}.\hskip 1em plus 0.5em minus 0.4em\relax IEEE, 2021, pp. 549--555.

\bibitem{a76}
P.~Li, S.~Gong, C.~Wang, and Y.~Fu, ``Ranking distance calibration for cross-domain few-shot learning,'' in \emph{Proceedings of the IEEE/CVF conference on computer vision and pattern recognition}, 2022, pp. 9099--9108.

\bibitem{a77}
Y.~Du, X.~Zhen, L.~Shao, and C.~G. Snoek, ``Hierarchical variational memory for few-shot learning across domains,'' \emph{arXiv preprint arXiv:2112.08181}, 2021.

\bibitem{a78}
L.~Sa, C.~Yu, X.~Ma, X.~Zhao, and T.~Xie, ``Attentive fine-grained recognition for cross-domain few-shot classification,'' \emph{Neural Computing and Applications}, vol.~34, no.~6, pp. 4733--4746, 2022.

\bibitem{a80}
M.~Yuan, C.~Cai, T.~Lu, Y.~Wu, Q.~Xu, and S.~Zhou, ``A novel forget-update module for few-shot domain generalization,'' \emph{Pattern Recognition}, vol. 129, p. 108704, 2022.

\bibitem{a81}
Q.~Liu and W.~Cao, ``Geometric algebra graph neural network for cross-domain few-shot classification,'' \emph{Applied Intelligence}, vol.~52, no.~11, pp. 12\,422--12\,435, 2022.

\bibitem{a82}
Q.~Zhang, Y.~Jiang, and Z.~Wen, ``Tacdfsl: Task adaptive cross domain few-shot learning,'' \emph{Symmetry}, vol.~14, no.~6, p. 1097, 2022.

\bibitem{a83}
W.~Yuan, T.~Ma, H.~Song, Y.~Xie, Z.~Zhang, and L.~Ma, ``Both comparison and induction are indispensable for cross-domain few-shot learning,'' in \emph{2021 IEEE International Conference on Multimedia and Expo (ICME)}.\hskip 1em plus 0.5em minus 0.4em\relax IEEE, 2021, pp. 1--6.

\bibitem{a84}
W.~Wang, L.~Duan, Y.~Wang, J.~Fan, and Z.~Zhang, ``Mmt: cross domain few-shot learning via meta-memory transfer,'' \emph{IEEE Transactions on Pattern Analysis and Machine Intelligence}, vol.~45, no.~12, pp. 15\,018--15\,035, 2023.

\bibitem{a85}
J.~Guan, M.~Zhang, and Z.~Lu, ``Large-scale cross-domain few-shot learning,'' in \emph{Proceedings of the Asian Conference on Computer Vision}, 2020.

\bibitem{a86}
W.~Chen, Z.~Zhang, W.~Wang, L.~Wang, Z.~Wang, and T.~Tan, ``Cross-domain cross-set few-shot learning via learning compact and aligned representations,'' in \emph{European Conference on Computer Vision}.\hskip 1em plus 0.5em minus 0.4em\relax Springer, 2022, pp. 383--399.

\bibitem{a87}
H.~Xu, L.~Liu, S.~Zhi, S.~Fu, Z.~Su, M.-M. Cheng, and Y.~Liu, ``Enhancing information maximization with distance-aware contrastive learning for source-free cross-domain few-shot learning,'' \emph{IEEE Transactions on Image Processing}, 2024.

\bibitem{a88}
Y.~Guo, R.~Du, Y.~Dong, T.~Hospedales, Y.-Z. Song, and Z.~Ma, ``Task-aware adaptive learning for cross-domain few-shot learning,'' in \emph{Proceedings of the IEEE/CVF International Conference on Computer Vision}, 2023, pp. 1590--1599.

\bibitem{a89}
N.~Paeedeh, M.~Pratama, M.~A. Ma’sum, W.~Mayer, Z.~Cao, and R.~Kowlczyk, ``Cross-domain few-shot learning via adaptive transformer networks,'' \emph{Knowledge-Based Systems}, vol. 288, p. 111458, 2024.

\bibitem{a90}
Y.~Zhao, T.~Zhang, J.~Li, and Y.~Tian, ``Dual adaptive representation alignment for cross-domain few-shot learning,'' \emph{IEEE Transactions on Pattern Analysis and Machine Intelligence}, vol.~45, no.~10, pp. 11\,720--11\,732, 2023.

\bibitem{a142}
V.~Papyan, X.~Han, and D.~L. Donoho, ``Prevalence of neural collapse during the terminal phase of deep learning training,'' \emph{Proceedings of the National Academy of Sciences}, vol. 117, no.~40, pp. 24\,652--24\,663, 2020.

\bibitem{a143}
O.~Vinyals, C.~Blundell, T.~Lillicrap, D.~Wierstra \emph{et~al.}, ``Matching networks for one shot learning,'' \emph{Advances in neural information processing systems}, vol.~29, 2016.

\bibitem{a144}
F.~Sung, Y.~Yang, L.~Zhang, T.~Xiang, P.~H. Torr, and T.~M. Hospedales, ``Learning to compare: Relation network for few-shot learning,'' in \emph{Proceedings of the IEEE conference on computer vision and pattern recognition}, 2018, pp. 1199--1208.

\bibitem{a145}
V.~Garcia and J.~Bruna, ``Few-shot learning with graph neural networks,'' \emph{arXiv preprint arXiv:1711.04043}, 2017.

\bibitem{a146}
H.~Wang and Z.-H. Deng, ``Cross-domain few-shot classification via adversarial task augmentation,'' \emph{arXiv preprint arXiv:2104.14385}, 2021.

\end{thebibliography}

\end{document}